\documentclass[12pt]{article}
\usepackage{graphicx} 
\usepackage{basicreq}
\usepackage{multirow}
\usepackage{amsmath,amssymb,amsfonts}
\usepackage{amsthm}
\usepackage{mathrsfs}
\usepackage[title]{appendix}
\usepackage{xcolor}
\usepackage{textcomp}
\usepackage{manyfoot}
\usepackage{booktabs}
\usepackage{algorithm}
\usepackage{algorithmicx}
\usepackage{algpseudocode}
\usepackage{listings}
\usepackage{multirow}
\usepackage{hhline}
\usepackage{rotating}
\usepackage{hyperref}
\usepackage{layout}
\usepackage{geometry}

\title{Role of single particle motility statistics on efficiency of targeted delivery of micro-robot swarms}
\author{Akshatha Jagadish, Manoj Varma }
\date{}

\begin{document}

\maketitle

\setlength{\oddsidemargin}{10pt}
\abstract{The study of dynamics of single active particles plays an important role in the development of artificial or hybrid micro-systems for bio-medical and other applications at micro-scale. Here, we utilize the results of these studies to better understand their implications for the specific application of drug delivery. We analyze the variations in the capture efficiency for different types of motion dynamics without inter-particle interactions and compare the results. We also discuss the reasons for the same and describe the specific parameters that affect the capture efficiency, which in turn helps in both hardware and control design of a micro-bot swarm system for drug delivery.}


\section{Introduction}\label{sec1}

Recent advancements in micro- and nano-fabrication technology is enabling rapid advances in the interdisciplinary field of targeted drug delivery systems \cite{Liu19}. The vision of targeted drug delivery system is captured well in the science fiction movie Fantastic Voyage (1966) in which a team of scientists travel to the site of infection in a shrunken submarine to treat a blood clot. While the current state of the art does not yet allow us to perform such a task, it is a future that many researchers are working towards \cite{Martinho11,Nava18}.

Currently, major Drug Delivery Systems (DDSs) are still oral or parenteral (intravenous, subcutaneous, and intramuscular) \cite{Srivastava18}. This has numerous problems such as the risk of adverse drug reactions, undesired toxic side effects, and low patient compliance, to name a few. These can be conceptually overcome with the help of targeted drug delivery systems. Paul Ehrlich put forward the seed for this thinking through his magic bullet concept \cite{Strebhardt08}: “Drugs that go straight to the targets” provide better pharmacokinetic properties which result in enhanced bioavailability of drugs by avoiding destruction by the immune system. Targeted DDSs also provide better pharmacodynamic properties by localised targeting and not affecting the healthy tissues and thus minimizing side effects. This is achieved by active manoeuvring in-vivo and controlled drug release. These properties result in a reduction in the intake frequency of drugs to maintain drug efficacy 

\renewcommand{\arraystretch}{1.5}
\begin{sidewaystable}
\caption{Examples of artifical microbots\label{tab:examples}}
\begin{tabular}{ |c|c|c|c|c|c| }
\hline
Sl. No. &	Name	& \begin{tabular}[c]{@{}c@{}}Type of \\ propulsion \end{tabular}	& \begin{tabular}[c]{@{}c@{}}Method of \\propulsion \end{tabular}&	Dimensions &	Reference\\ [0.5ex]
\hline
\hline
1 &	Janus particles &	Internal &	Chemical	& 50-500nm &	\cite{Ma15}\\
\hline
2 &	Micro helices &	External &	Magnetic &	20um &	\cite{Servant15} \\
\hline
3 &	Micro-tubes &	Internal	& Chemical &	15x5um &	\cite{Li16} \\
\hline
\multirow{2}{*}{4}& \multirow{2}{8em}{Rod shaped micro particles} & External & Acoustic & 2x0.3um & \cite{Wang12} \\ \hhline{~~----}
& \multirow{2}{*}{} & Internal & Chemical & 2x0.37um & \cite{Paxton04} \\ 
\hline
5 &	Polymeric vesicles 	& Internal &	Chemical &	0.1um &	\cite{Joseph17} \\
\hline
6 &	Cilia/ flexible tail &	External &	Magnetic &	\textless10mm, 220x60um &	\cite{Lum16,Kim16}\\
\hline
7 &	Micro-gears	& External	& Light	& 16um dia &	\cite{Maggi15}\\
\hline
8 &	Bent microrods &	Internal &	Chemical &	30um avg x 2-6um &	\cite{Rao18}\\
\hline
9 &	Hollow microspheres &	External &	Photothermal &	\begin{tabular}[c]{@{}c@{}}0.5-2um \\(mean = 0.876um) \end{tabular}	 & \cite{Zha13} \\
\hline
10 & \begin{tabular}[c]{@{}c@{}}Peanut shaped  \\ micro-particles\end{tabular}	 & External &	Magnetic &	3umx2um &	\cite{Xie19}\\
\hline
11 &	L-shaped micro structures &	External &	Photothermal & 	Long arm: 9um, short arm: 6um &	\cite{Kummel13}\\
\hline
\end{tabular}
\end{sidewaystable}

Development of such targeted DDSs has attracted the attention of not just biological and chemical fields but also the engineering fields such as nano-technology due to its ability to manipulate objects at micro- and nano-scale. Thus, there have been several approaches for designing ‘microbots’ that can act as a magic bullet or a shrunken submarine \cite{Ceylan17}.  A few of these examples are shown below in Table \ref{tab:examples}.

In Table \ref{tab:examples}, the type of propulsion is either external or internal, where external propulsion indicates that the entire swarm or group of microbots is globally powered and remotely actuated and guided, whereas internal propulsion indicates that each microbot in the swarm is autonomous (self-contained) and thus self-propelled. The external propulsion mechanism is simple in its implementation but has limited capabilities. The internal propulsion mechanism on the other hand is complex but once implemented will have emergent intelligence. A few patterns that emerge in the above table are as follows. Internally powered microbots are usually chemically propelled. Chemical propulsion occurs by the reaction of microbot’s material with the surrounding fluid, thus converting chemical energy into kinetic energy. It can be observed that microbots with size below 3 microns (suitable for bio-applications) are either magnetically or chemically powered. 

Whichever the implementation technique, an effective DDS is the one that can achieve efficient navigation towards the target region. Here, efficient navigation is an umbrella term that covers various aspects of optimal behaviour such as being the quickest to reach the target or consuming least power or having the largest capture efficiency and so on. We focus on the aspect of capture efficiency, where refers to the fraction of microbots reaching a specific “target” region, representing an organ to which drugs may need to be delivered. An ideal DDS would possess 100\% or unit (normalized to 1) capture efficiency. However, this is not possible in general because the microbots do not follow an exact path unlike their macro-scale counterparts because of the dominance of random Brownian motion at the micro-scale. This renders them hard to be controlled. The stochastic dynamics of the swarm of these delivery agents (microbots) play an important role in determining the efficiency of the DDS.

There has been a lot of research aiming to describe the dynamics of such microbots in the context of natural and artificial active matter systems \cite{Bechinger16,Fodor18,Shaebani20,Hauser15}. In addition to trying to understand the behaviour of active systems, the dynamical models reported in literature also consider the effects of surroundings, collective phenomena, boundaries, and various other environmental and emergent effects. The most generic of these are Active Brownian Particle (ABP), Run and Tumble Particle (RTP), Chiral ABP and Passive Brownian Particle (PBP). Experimentally observed dynamics of microbots can generally be well described by one of these generic models \cite{Bechinger16}. Differences in fabrication, mode of propulsion and so on lead to difference in dynamical behaviour at the single agent (microbot) level \cite{Vutukuri16}. For instance, the behaviour of individual microbot may be much better described by a chiral ABP model as opposed to an ABP model (Examples 7 and 8 in table \ref{tab:examples}). To the best of our knowledge, the role of individual motility dynamics (statistics) on the overall capture efficiency of a swarm of microbots has not been reported in existing literature. Therefore, in this work, we characterize the efficiency of a DDS, an artificial, micro active matter system by examining the four generic motility models mentioned above. We first describe the mathematical details of the four generic models. Then we compare them using the parameter ‘capture efficiency’, which denotes the fraction of microbots that were successful in reaching the target. 

\section{Description of Models}
In this section, we introduce the mathematical details of the four models by providing the dynamical equations and qualitative behaviour. The density map or the position PDF is a good measure to compare the four models and has been studied extensively \cite{Fodor18,Wagner17,Shaebani20,Kurzthaler16,Malakar17}. 

\subsection{Active Brownian Particle (ABP)}
This is an extensively used model to describe a non-interacting self-propelled particle. The ABP model was developed to describe self-phoretic colloids and non-tumbling E-coli bacteria \cite{Bechinger16,Basu18,Fodor18}. These particles move with a constant velocity v but their orientation gradually changes due to the rotational diffusion co-efficient D\textsubscript{R}. Here, the particle dynamics in 2 dimensions is modelled as in equation \ref{eqn:ABP_xyphi}.

\begin{subequations} \label{eqn:ABP_xyphi}
    \begin{align}
        \dot{x} &= v\cos\phi + \sqrt{2D_T}\xi_{x} \label{eqn:ABP_x}\\
        \dot{y} &= v\sin\phi + \sqrt{2D_T}\xi_{y} \label{eqn:ABP_y}\\
        \dot{\phi} &= \sqrt{2D_R}\xi_{\phi} \label{eqn:ABP_phi}
    \end{align}
\end{subequations}

($x$,$y$) denotes the position of the particle  ($\dot{x}$, $\dot{y}$ are used to update $x$,$y$ over time), $\phi$ denotes the orientation, $D_T$ is the translational diffusion co-efficient, $D_R$ is the rotational diffusion co-efficient, $\xi$ denotes the Gaussian white noise in equation \ref{eqn:ABP_xyphi}. 

The mean square displacement (MSD) of ABP over time is shown in equation \ref{eqn:ABP_MSD}.
 
\begin{equation}
    \text{MSD} = 4D_Tt + \frac{2v^2t}{D_R} - \frac{2v^2(1-e^{-D_Rt})}{D^2_R}
    \label{eqn:ABP_MSD} 
\end{equation}

\subsection{Run and Tumble Particle (RTP)}
This is a well-known model developed to describe the dynamics of predominantly natural micro-organisms like bacteria \cite{Cates12}. The particle dynamics involves periods of runs and tumbles. In the run state, the particle moves with a constant velocity for some duration, and in the tumble state, it changes its orientation while being in the same location. Here, the particle dynamics in 2 dimensions is modelled as in equation \ref{eqn:RTP_xyphi}.

\begin{subequations} \label{eqn:RTP_xyphi}
    \begin{align}
        \dot{x} &= v\cos\phi + \sqrt{2D_T}\xi_{x} \label{eqn:RTP_x}\\
        \dot{y} &= v\sin\phi + \sqrt{2D_T}\xi_{y} \label{eqn:RTP_y}\\
        \dot{\phi}(t) &= \sum_{i}{\Delta \phi_i}\delta(t-T_i) \label{eqn:RTP_phi}
    \end{align}
\end{subequations}

The above change in $\phi$ occurs at a rate $\alpha$ called the tumble rate. The MSD of RTP is shown in equation \ref{eqn:RTP_MSD} \cite{Santra20}.

\begin{equation}
    \text{MSD} = 4D_Tt + \frac{2v^2t}{\alpha} - \frac{2v^2(1-e^{-\alpha t})}{\alpha^2}
    \label{eqn:RTP_MSD} 
\end{equation}

\subsection{Chiral ABP}
These particles are similar to ABPs except for an addition of angular velocity $\omega$ in their dynamics as described in the equation \ref{eqn:Chi_xyphi}. 

\begin{subequations} \label{eqn:Chi_xyphi}
    \begin{align}
        \dot{x} &= v\cos\phi + \sqrt{2D_T}\xi_{x} \label{eqn:Chi_x}\\
        \dot{y} &= v\sin\phi + \sqrt{2D_T}\xi_{y} \label{eqn:Chi_y}\\
        \dot{\phi} &= \omega + \sqrt{2D_R}\xi_{\phi} \label{eqn:Chi_phi}
    \end{align}
\end{subequations}

The chirality arises due to a small deviation in the symmetry of the microbot or in the propulsion mechanism and is observed a lot in nature \cite{Bechinger16}. This deviation is taken care by the additional parameter $\omega$ which could play an important role in the capture efficiency of targeted DDS as can be seen later.

MSD of the Chiral ABP can be calculated as shown in equation \ref{eqn:Chi_MSD} empirically derived from the MSD of vescicles filled with Chiral ABP particles \cite{Chen17}. 

\begin{equation}
    \begin{aligned}
    \label{eqn:Chi_MSD}
        \text{MSD}  &= 4D_Tt + \frac{2v^2D_Rt}{D_R^2+\omega^2}\\ 
            &\qquad + \frac{2v^2(e^{-D_Rt}\cos (\omega t+\phi_0)-\cos(\phi_0))}{D^2_R+\omega^2}
    \\[4pt]
        \cos \phi_0 &= \frac{D^2_R-\omega^2}{D^2_R+\omega^2}
    \end{aligned}
\end{equation}

\subsection{Passive Brownian Particle (PBP)}
These are simple non-propelled particles undergoing Brownian motion and their dynamics is described in equation \ref{eqn:PBP_xyphi}. 

\begin{subequations} \label{eqn:PBP_xyphi}
    \begin{align}
        \dot{x} &=  \sqrt{2D_T}\xi_{x} \label{eqn:PBP_x}\\
        \dot{y} &=  \sqrt{2D_T}\xi_{y} \label{eqn:PBP_y}\\
        \dot{\phi} &= \sqrt{2D_R}\xi_{\phi} \label{eqn:PBP_phi}
    \end{align}
\end{subequations}

The active or propelled particles (ABP, PBP, Chiral ABP) definitely have a better performance when compared to PBP [1]. Hence, in our paper, we focus on  the relative performance among the different active particles.

Examination of equations \ref{eqn:ABP_xyphi}-\ref{eqn:PBP_xyphi} clearly shows that at long time-scales, i.e., $t >> 1/D_R$, all the four models can be mapped to each other. For instance, there have been studies showing the equivalence of ABPs and RTPs at long time scales \cite{Cates13, Solon15}. In addition, both types of particles accumulate at the boundaries in confined geometries \cite{Khatami16,Malakar17,Wensink08}. Despite many such similarities, there are differences observed between the two types of particles at short time scales \cite{Khatami16}.

\begin{figure}[t]
    \centering
    \includegraphics[width=\linewidth]{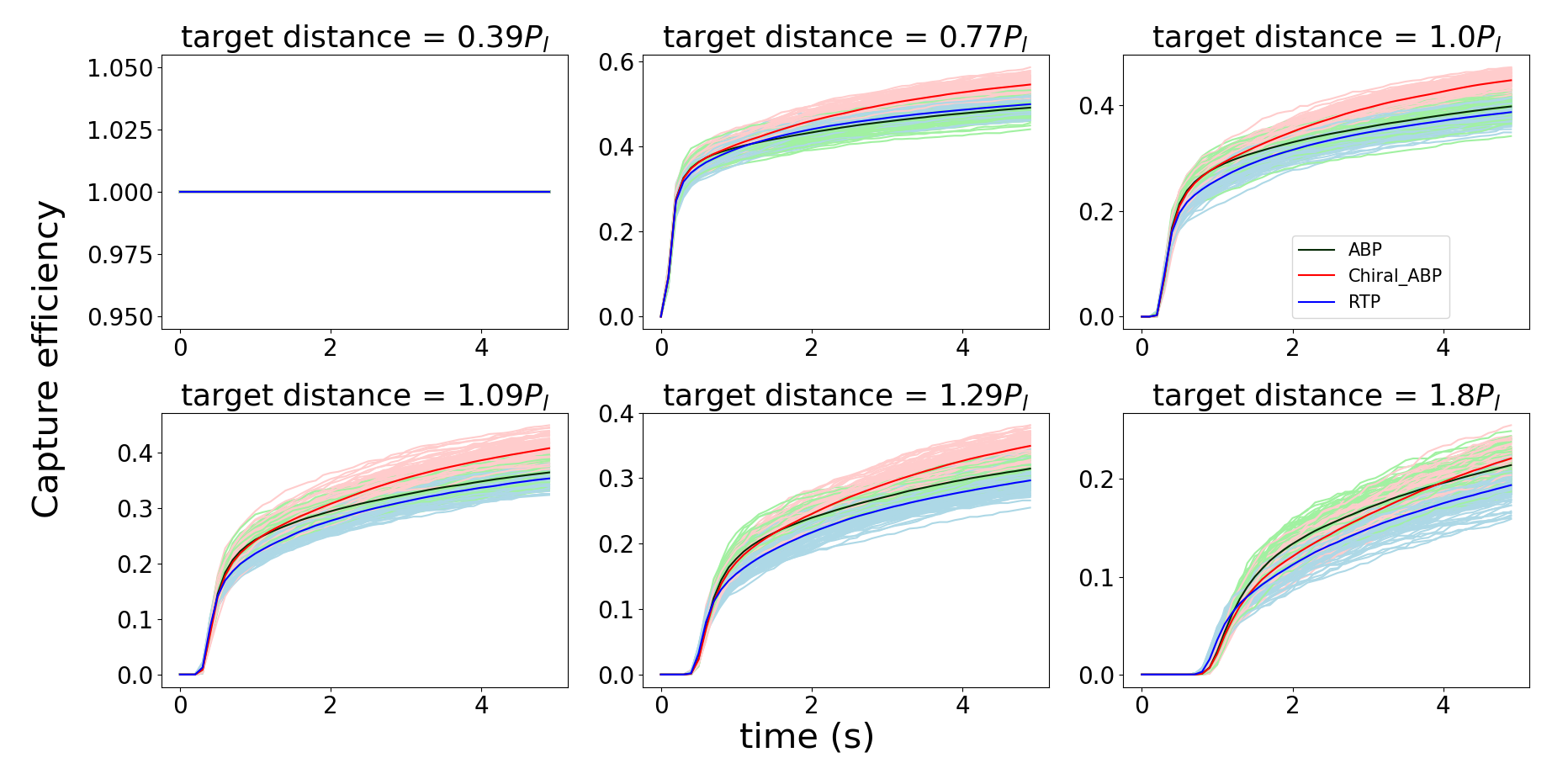}
    \caption{Capture efficiency for 1000 particles}
    \label{fig:100_sims_cap_eff}
\end{figure}

\begin{figure}[t]
    \centering
    \includegraphics[width=\linewidth]{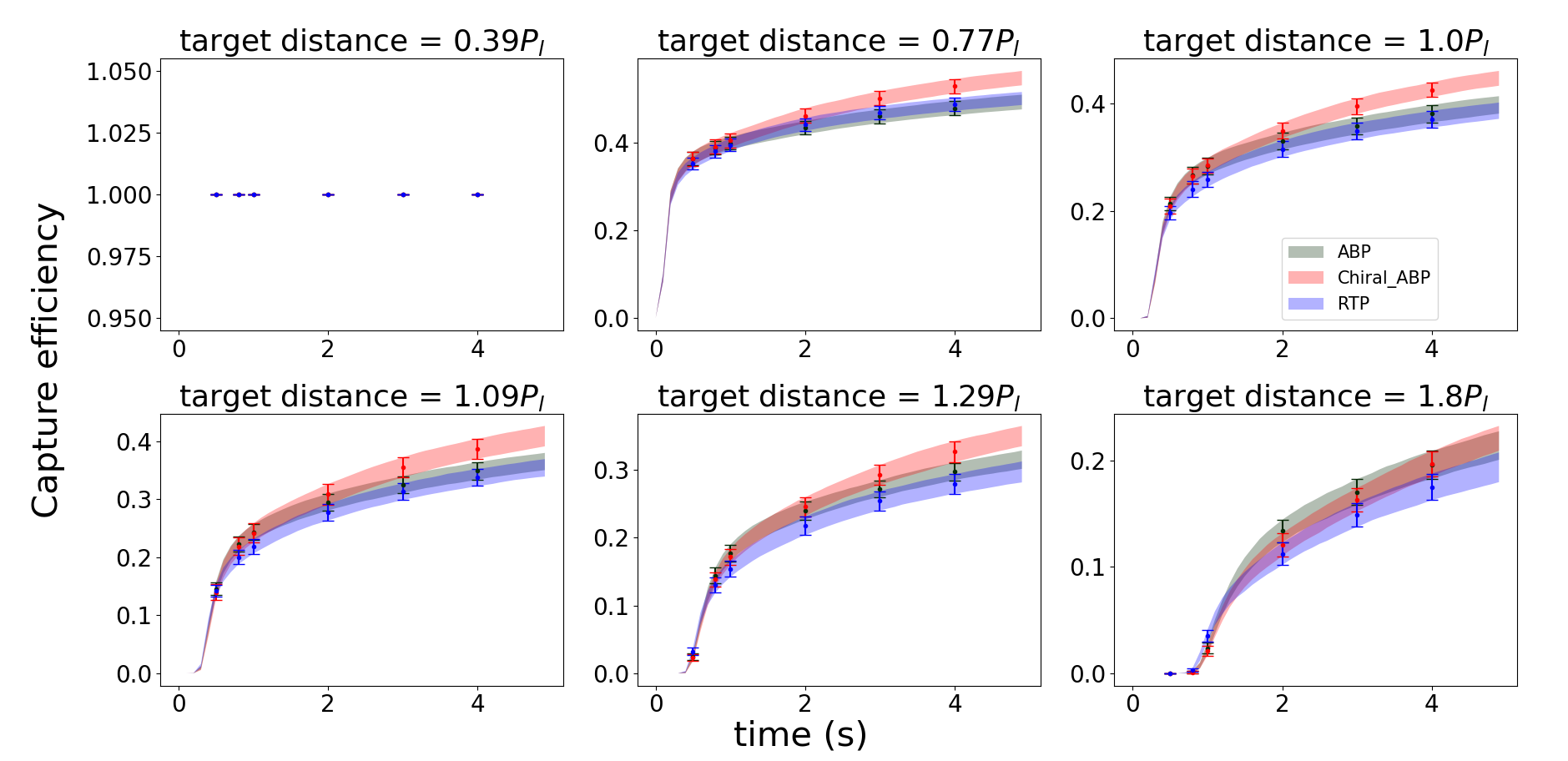}
    \caption{Error bar of capture efficiency}
    \label{fig:100_sims_error_bar}
\end{figure}

\begin{figure}[t]
    \centering
    \includegraphics[width=\linewidth]{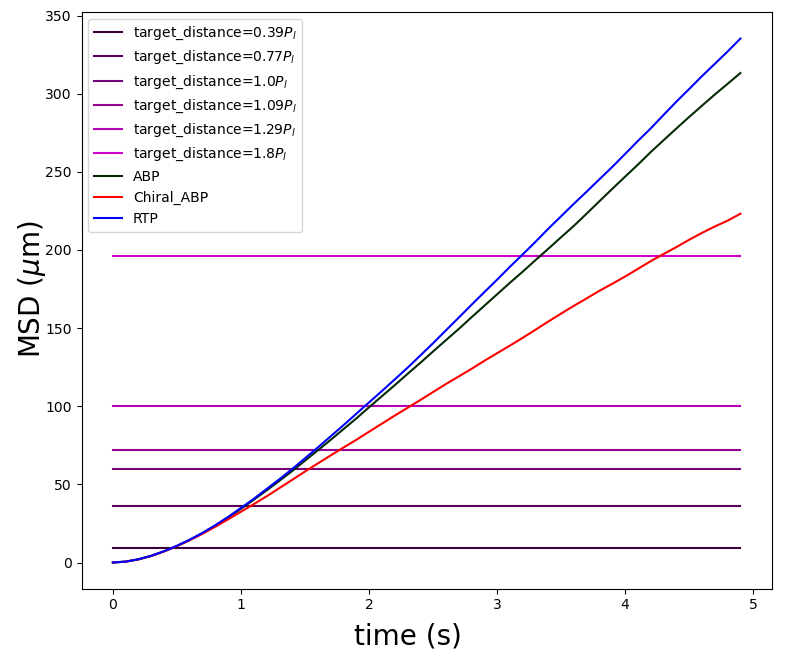}
    \caption{Mean Squared Displacement (MSD) plot}
    \label{fig:MSD}
\end{figure}

\section{Methods}
We simulated ABP, RTP and Chiral ABP in 2D using the finite difference equations of the dynamic equations of their position described in the equations \ref{eqn:ABP_xyphi}, \ref{eqn:RTP_xyphi} and \ref{eqn:Chi_xyphi} \cite{Callegari19}. We performed the simulations using a custom Python script (version 3.8.5). 

Active particles at the micro-scale have a noisy movement which is quantified by the diffusion constants, which depend on the shape and size of the particle. $D_R$ and $D_T$ are the rotational and translational diffusion constants, respectively and are determined as shown in equation \ref{eqn:D} for spherical (circular in 2D) particles. In our simulations, the radius of the particle $R=0.5\mu m$. $\alpha$ for RTP is kept equal to $D_R$ to check the difference in capture efficiency when ABP and RTP are equivalent in 2D \cite{Cates13}. $\omega$ for Chiral ABP is fixed at $1rads^{-1}$. 

\begin{subequations} \label{eqn:D}
    \begin{align}
        D_T &=  \frac{k_Bt}{6\pi \eta R} \label{eqn:D_T}\\
        D_R &=  \frac{k_Bt}{8\pi \eta R^3} \label{eqn:D_R}
    \end{align}
\end{subequations}

In equation \ref{eqn:D}, $k_B$ denotes the Boltzmann constant and is of value $1.38064852 \times10^{-23} m^2 kg s^{-2} K^{-1}$ ,$T$ is the absolute temperature in $Kelvin$, $\eta$ is the fluid viscosity (we consider water as the fluid here, hence $= 1.0016 \times 10^{-3} Pa s$). For $R=0.5\mu m$, $D_T$ and $D_R$ are calculated to be $\mu m^2s^{-1}$ and $rad^2s^{-1}$ respectively.

The target for our simulated DDS is designed to be circular and absorbing in nature, meaning a particle entering this region is stuck there and is said to be “delivered”. The target is taken to be of radius $a (=5\mu m)$ and is located at a distance $l$ from the point of the initial location of the particles. The capture efficiency is measured at each time instant $\delta t$ by noting the number of particles making it to the target. 

The code is available in the link, \href{https://github.com/Akshatha-Jagadish/micro-robotic-simulations}{code} for 2d and 3d simulations of ABP, RTP and Chiral ABP.

\begin{figure}[t]
    \centering
    \includegraphics[width=\linewidth]{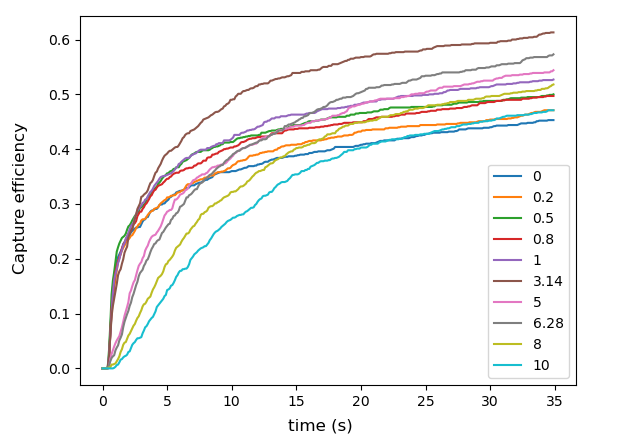}
    \caption{Effect of $\omega$ on Chiral ABP capture efficiency}
    \label{fig:effect_of_w}
\end{figure}

\begin{figure}[t]
    \centering
    \includegraphics[width=\linewidth]{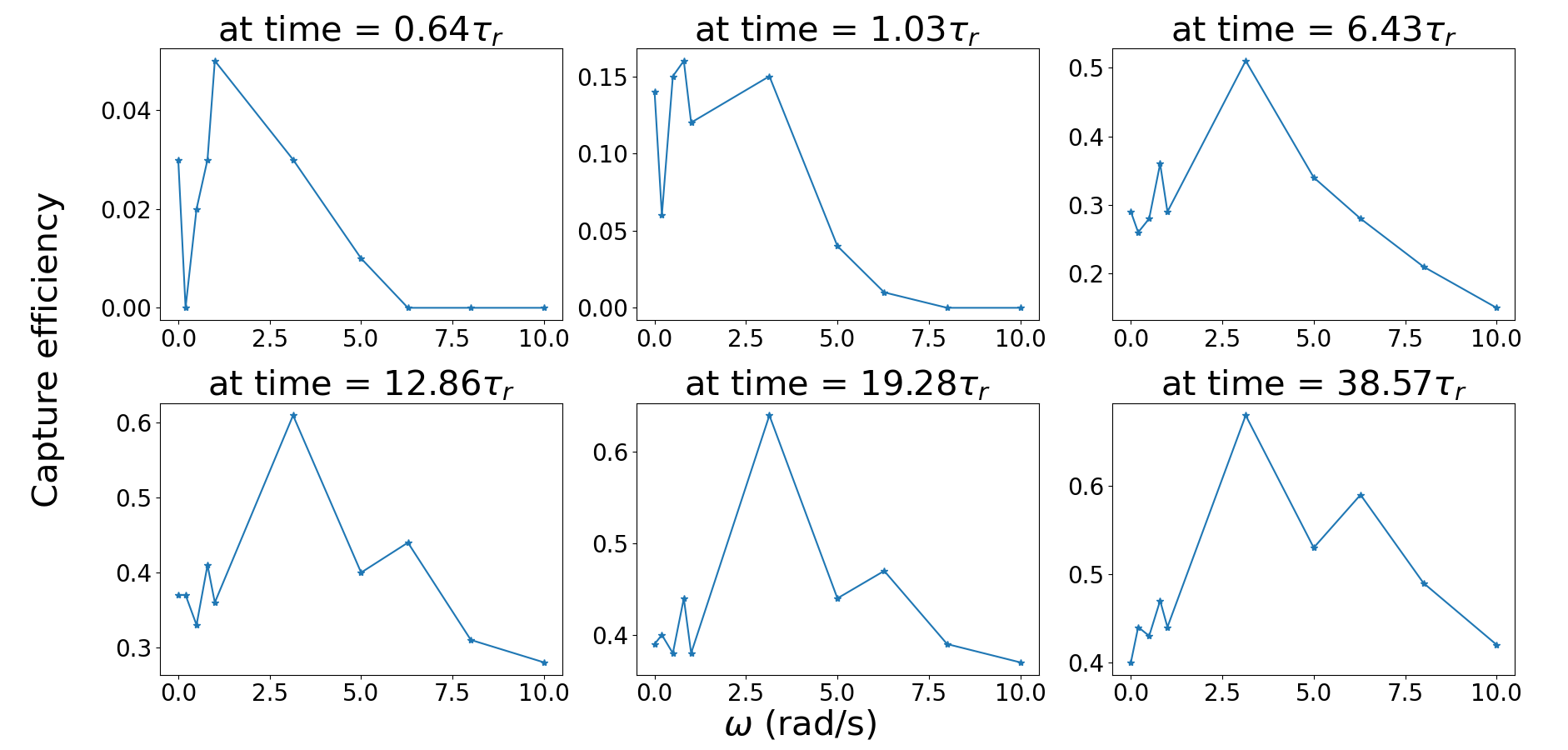}
    \caption{Effect of $\omega$ on Chiral ABP capture efficiency at specific times}
    \label{fig:effect_of_w_times}
\end{figure}

\section{Results}
In literature, the quantity persistence length described by $P_l = v/D_R$ gives a measure of how far an active particle maintains the initial orientation on average; and the quantity $\tau_R = 1/D_R = 0.778s$ gives a measure of how long an active particle maintains the initial orientation on average. We thus simulated the trajectory of 1000 particles and ran the simulation 100 times to obtain the statistics of capture efficiency for range of target distance $l$ both less and more than $P_l = 7.778\mu m$ for $v=10\mu ms^{-1}$ and for simulation time $t=5s >> \tau_R$.

To determine the effect of motility type (i.e., ABP, RTP and chiral ABP) on the capture efficiency, we plotted the simulated capture efficiencies for the different motility types for different target-distances, as shown in Figure \ref{fig:100_sims_cap_eff}. We see that the curves for the different motility types overlap for small or large target distances. However, the chiral ABP provides higher capture efficiency for intermediate target distance compared to the other two motility types. The same observation is illustrated in Figure \ref{fig:100_sims_error_bar} where we have plotted the mean and standard deviation of capture efficiency for the different motility types.

To understand the particles' behaviour, we compare the time evolution of the Mean Squared Displacement (MSD) for the different motility types as shown in figure \ref{fig:MSD}. As pointed out before, at time scales longer than $1/D_R$; the MSD for RTP and ABP particles can be mapped into each other. 

In figure \ref{fig:100_sims_cap_eff}, we observe that the Chiral ABP has a higher capture efficiency when compared to RTP and ABP at target distances closer to persistence length. This is because the effective diffusion constant of a Chiral ABP is always smaller when compared to that of either ABP or RTP (kept equal) which increases the approximate time spent “near” the target area (roughly $a^2/D$). This is also evidenced by the MSD plot in figure 3, where chiral ABP has lower MSD and, therefore, smaller effective $D$.

We also note that at higher target distances, the capture efficiency of ABP gradually improves and matches with that of Chiral ABP as observed in figure \ref{fig:100_sims_error_bar}. This is because, for targets that are far away, the ABP's better initial directed motion makes up for the advantage of Chiral ABP's behaviour.

To understand the effect of angular velocity, we plot capture efficiency of Chiral ABPs by repeating the simulations for different values of $\omega$ as shown in figure \ref{fig:effect_of_w}. Here, we observed a particular $\omega$ for which the capture efficiency was maximum. Also, this observation was prominent at times $>= \tau_r$ (figure \ref{fig:effect_of_w_times}). This phenomenon is explained as follows. For a fixed target distance, capture efficiency reaches saturation at time $t_s = l^2/D$. So for a fixed time $t$, the $\omega$ corresponding to $l^2/t$ will provide the maximum capture efficiency as shown in the figure on $\omega$ variation.

We also simulated the ABP, RTP and Chiral ABP in 3D and plotted the capture efficiency as shown in figure 6. Here, we observed that the behaviour in 2D is also reflected in 3D. 

\begin{figure}[t]
    \centering
    \includegraphics[width=\linewidth]{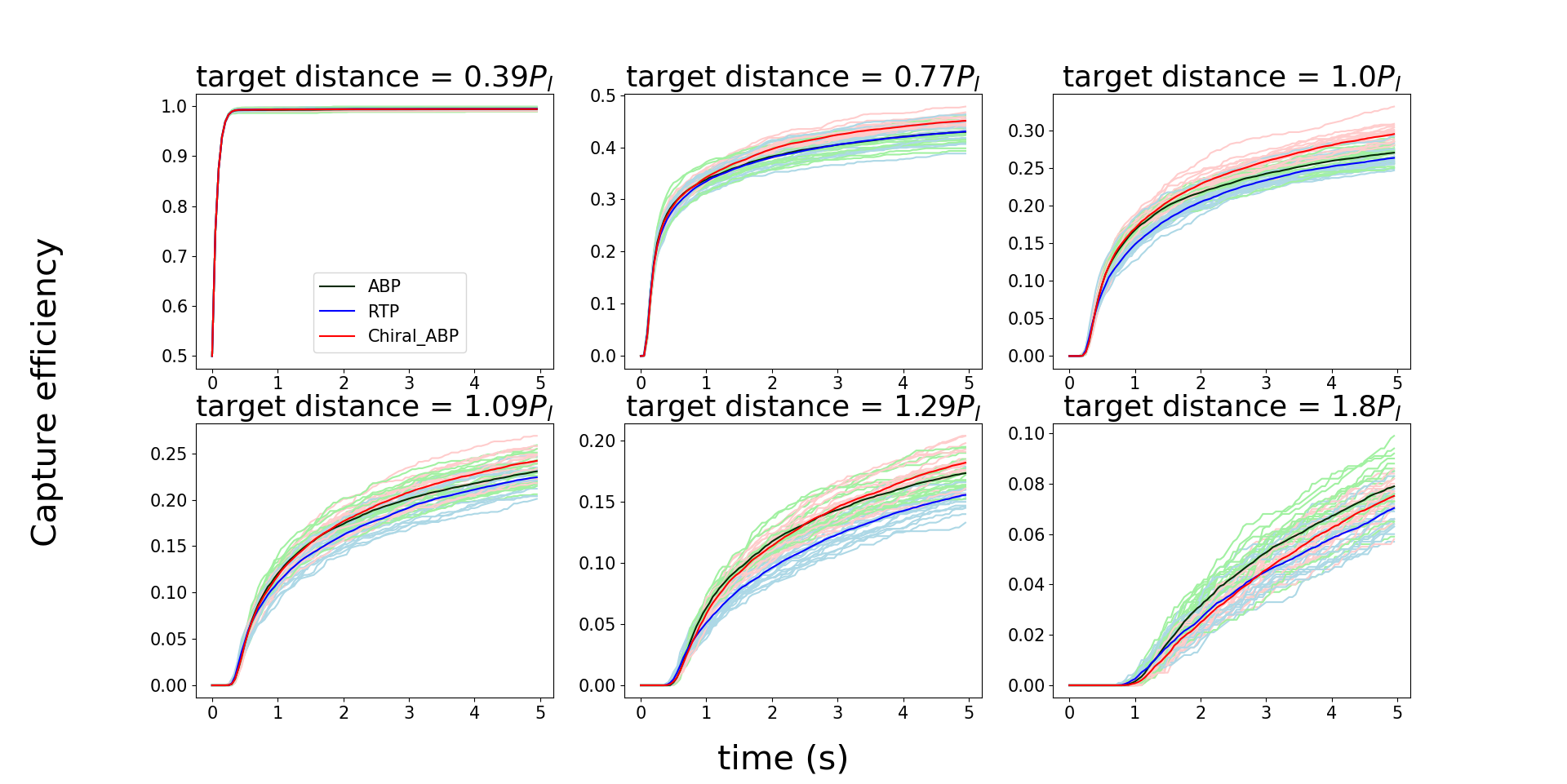}
    \caption{Capture efficiency for 1000 particles moving in 3 dimensions}
    \label{fig:3d_cap_eff}
\end{figure}

\section{Conclusion}
The following conclusions were drawn from the simulation of active particles with different motility characteristics.
\begin{itemize}
\item{The particle having MSD that reaches the target latest has the best capture efficiency.}
\item{Effect of w for Chiral ABP: there is an optimum w for a particle with certain specifications.}
\item{When target distance is close to the persistence length, RTP and ABP have equal performances while Chiral ABP fares better than both. When target distance is much higher than persistence length, ABP and Chiral ABP have better and similar performances when compared to RTP.}
\end{itemize}
This work presents the impact of a single-particle level property, namely the motility statistics with no inter-particle interactions, on a macroscopic swarm-level parameter, namely delivery or capture efficiency. As such the observations described here will be helpful to tune single particle-level characteristics to maximize swarm-level performance.

\bibliographystyle{IEEEtran}

\bibliography{sn-article}

\end{document}